\documentclass[sigconf]{acmart}

\usepackage{import}
\usepackage{algorithmic}
\usepackage{algorithm}

\usepackage{textcomp}
\usepackage{multirow}
\usepackage{booktabs}
\usepackage{amsmath,xparse,mleftright}
\usepackage{balance}

\AtBeginDocument{%
  }

\copyrightyear{2023} 
\acmYear{2023} 
\setcopyright{acmlicensed}
\acmConference[MM '23]{Proceedings of the 31st ACM International Conference on Multimedia}{October 29-November 3, 2023}{Ottawa, ON, Canada}
\acmBooktitle{Proceedings of the 31st ACM International Conference on
 Multimedia (MM '23), October 29-November 3, 2023, Ottawa, ON, Canada}
\acmPrice{15.00}
\acmDOI{10.1145/3581783.3612201}
\acmISBN{979-8-4007-0108-5/23/10}

\begin{document}

\title{Toward Zero-shot Character Recognition: A Gold Standard Dataset with Radical-level Annotations}



\author{Xiaolei Diao}
\email{xiaolei_diao@163.com}
\orcid{0000-0002-3269-8103}
\affiliation{%
  \institution{College of Computer Science and Technology, Jilin University}
  \country{}
}
\affiliation{%
  \institution{DISI, University of Trento}
  \country{}
}

\author{Daqian Shi}
\email{daqians123@163.com}
\orcid{0000-0003-2183-1957}
\affiliation{%
  \institution{College of Computer Science and Technology, Jilin University}
  \country{}
}
\affiliation{%
  \institution{DISI, University of Trento}
  \country{}
}

\author{Jian Li}
\email{lijianjlu@126.com}
\orcid{0000-0002-0065-4333}
\affiliation{%
  \institution{College of Computer Science and Technology, Jilin University}
  \country{}
}

\author{Lida Shi}
\email{shild21@mails.jlu.edu.cn}
\orcid{0000-0001-5011-6931}
\affiliation{%
  \institution{School of Artificial Intelligence, Jilin University}
  \country{}
}

\author{Mingzhe Yue}
\email{yuemingzhe.jlu@gmail.com}
\orcid{0009-0008-9982-2026}
\affiliation{%
  \institution{College of Computer Science and Technology, Jilin University}
  \country{}
}

\author{Ruihua Qi}
\email{qirh20@mails.jlu.edu.cn}
\orcid{0000-0002-0554-6040}
\affiliation{%
  \institution{School of Archaeology, Jilin University}
  \country{}
}

\author{Chuntao Li}
\email{lct33@jlu.edu.cn}
\orcid{0000-0001-9836-1493}
\affiliation{%
  \institution{School of Archaeology, Jilin University}
  \country{}
}
\authornote{Corresponding authors.}

\author{Hao Xu}
\email{xuhao@jlu.edu.cn}
\orcid{0000-0001-8474-0767}
\affiliation{%
  \institution{College of Computer Science and Technology, Jilin University}
  \country{}
}
\authornotemark[1]

\renewcommand{\shortauthors}{Xiaolei Diao et al.}


\begin{abstract}
Optical character recognition (OCR) methods have been applied to diverse tasks, e.g., street view text recognition and document analysis. Recently, zero-shot OCR has piqued the interest of the research community because it considers a practical OCR scenario with unbalanced data distribution. However, there is a lack of benchmarks for evaluating such zero-shot methods that apply a divide-and-conquer recognition strategy by decomposing characters into radicals. Meanwhile, radical recognition, as another important OCR task, also lacks radical-level annotation for model training. In this paper, we construct an ancient Chinese character image dataset that contains both radical-level and character-level annotations to satisfy the requirements of the above-mentioned methods, namely, ACCID, where radical-level annotations include radical categories, radical locations, and structural relations. To increase the adaptability of ACCID, we propose a splicing-based synthetic character algorithm to augment the training samples and apply an image denoising method to improve the image quality. By introducing character decomposition and recombination, we propose a baseline method for zero-shot OCR. The experimental results demonstrate the validity of ACCID and the baseline model quantitatively and qualitatively.
\end{abstract}





\maketitle

\section{Introduction}
\label{sec:1}
\begin{figure}[!t]
	\centering
	\setlength{\abovecaptionskip}{0pt}%
    \setlength{\belowcaptionskip}{0pt}%
	\includegraphics[width=1\linewidth]{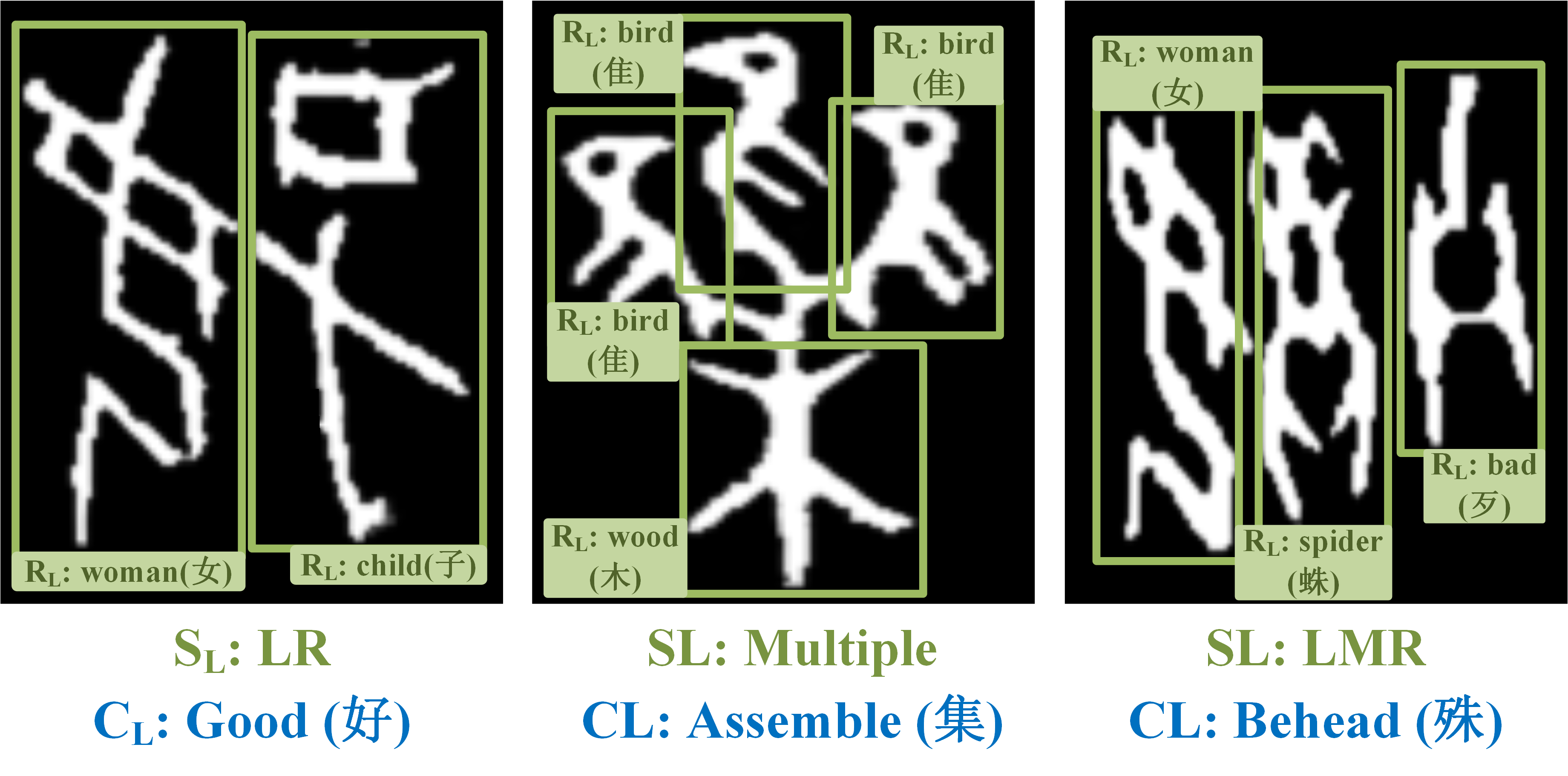}
	\caption{Examples of annotated images in the ACCID, on the character-level and radical-level. \label{fig:accid}}
\end{figure}

Optical character recognition (OCR) is one of the critical techniques for managing, restoring, and utilizing existing character image resources. Specifically, the concept of character images broadly includes different carriers (e.g., documents, bronzes, and street-view texts) of various languages (e.g., English and Chinese) and fonts (e.g., handwritten and printed texts) \cite{shi2022rcrn}. Automatic OCR methods are needed to deal with large-scale and domain-specific character images. With the development of deep learning (DL) in recent years, OCR methods have achieved significant improvements on diverse tasks \cite{liu2021oracle}, e.g., document character extraction and historical character recognition.

General DL-based methods pursue a large number of training samples with balanced distribution, while such an ideal scenario is difficult to satisfy in the real world. In the OCR area, we also observe the same unbalanced sample issue due to the differences in character usage frequency \cite{diao2023rzcr}, which significantly affects recognition performance in practice. As a result, researchers attempt to introduce radicals in their OCR methods to alleviate the effect of unbalanced distribution, where radicals are the main component for forming characters and presenting character semantics. The main idea comes from that compared with characters, the distribution of radicals is more balanced since the same radical can be shared by different characters. As shown in Figure ~\ref{fig:accid}, characters are composed of radicals in certain structures, where radicals are highlighted in green boxes and we can find the radical ``women" is shared by the first and third characters. Meanwhile, the pre-training model based on radicals is also of interest to researchers, since radicals include huge semantic information that is useful to various downstream tasks. Some researchers classify radicals in character images, considering radical recognition as a standard multi-class classification task \cite{wang2017radical}.

According to the current studies on radicals, we found that effective utilization of radical information will be beneficial to character recognition. In recent years, researchers explored character decomposition and learned from decomposed elements to achieve zero-shot character recognition \cite{zhang2018radical, zhang2020radical}. These radical-based OCR methods have the ability to recognize few-sample character categories or even unseen character categories by applying zero-shot learning. Current zero-shot methods usually attempt to learn from the character images and extract their radical sequences. However, such methods do not well-performed in practice since the extracted radical sequence which is a kind of character-level annotation, needs to be strictly matched with the target label, where a minor mistake can lead to failed results. Thus, annotated character image dataset with concrete radical-level information is needed for training and evaluating zero-shot OCR methods.

In this study, we construct an ancient Chinese character image dataset that contains both radical-level and character-level annotations, namely ACCID, as a benchmark for zero-shot OCR methods. Firstly, we propose a character decomposition dictionary to store the correspondences between characters and radicals, where 2,892 character categories and 595 radical categories are included. Then, we collect ancient Chinese character images from \cite{wu2012} and annotate the characters, radicals, and structures following the character decomposition dictionary \cite{chi2022zinet}. In ACCID, character images have both character-level and radical-level annotations, as shown in Figure ~\ref{fig:accid}, where character categories are marked in blue and radical-level annotations, including radical categories, radical coordinates, and structure categories are highlighted in green, respectively. Furthermore, we introduce an effective splicing-based synthetic character algorithm to increase radical-level annotations without additional human efforts, which can further augment the training data to support DL-based methods to recognize characters and radicals. Besides, we propose a baseline method based on character decomposition and reconstruction to verify the effectiveness of zero-shot character recognition. Specifically, we first recognize the radicals and structures of characters, and then perform reasoning based on a character dictionary to obtain character categories by utilizing the identified radical-level information. Experimental results prove the validity of our proposed ACCID and baseline method.

The contributions of our study are summarized as follows:
\vspace{-0.1cm}
\begin{itemize}
    \item We propose a novel ancient Chinese character image dataset ACCID with both radical-level and character-level annotations, for supporting the training of zero-shot character recognition and radical recognition methods. 

    \item In the construction of ACCID, we introduce a character decomposition dictionary to guide radical-level annotations of character images and an effective splicing-based synthetic character algorithm to augment the training samples. 

     \item We propose a baseline method based on character decomposition and recombination for zero-shot character recognition. Experimental results demonstrate the advantages of the zero-shot method in few-sample character datasets.

    \item ACCID is evaluated explicitly by experts and implicitly by existing deep learning-based recognition methods. The experimental results demonstrate the validity of ACCID. 
\end{itemize}
\vspace{-0.1cm}

It is worth mentioning that the character images collected in ACCID are rubbings of unearthed oracle bone inscriptions (OBI) \cite{oracle}, which is one of the oldest characters in the world. Due to the scarcity of ancient characters and the unbalance character usage frequency, a large number of character categories have insufficient samples, which poses great challenges to recognition. The study of oracle bone inscriptions is of vital importance in exploring and understanding ancient culture \cite{huang2019obc306}. Thus, the oracle bone character recognition task offers valuable aid in the preservation and investigation of traditional culture.



\section{Related Work}
\label{sec:2}
\subsection{Character Recognition}
Character recognition is one of the fundamental tasks in the computer vision area. In early research, OCR methods tended to utilize pre-set features from character images by filter-based techniques \cite{su2003novel,heutte1998structural, uchida2003eigen}. Recently, OCR methods achieve great improvements since the development of DL. Convolutional neural networks (CNN) are applied to extract deep features from the character images for better recognition performance \cite{cirecsan2015multi, zhong2015high}. Feng et al. \cite{feng2015recognition} propose a fuzzy character recognition model based on a statistical analysis of context and the Hopfield network. Li et al. \cite{li2018building} propose a cascaded model in a single CNN with global weighted average pooling to achieve higher recognition accuracy with a limited number of parameters. The performance of these DL-based OCR methods relies on the data quality of the character datasets with character-level annotations. However, the difference in the usage frequency of characters and the huge number of character categories limit model training in practice. Moreover, in real-world scenarios, there are character categories not included in the training set \cite{wang2019radical}, which also challenge these OCR methods.

In order to address the problems mentioned above, researchers attempt to exploit radicals for East Asian character recognition. Wang et al. \cite{wang2018denseran} first propose a radical analysis network with densely connected architecture to learn radicals and two-dimensional structures of characters from the character images. They analyze the radicals and structures through an attention-based decoder to achieve the recognition of unseen handwritten characters. Zhang et al. \cite{zhang2018radical} propose an encoder-decoder structure with a spatial attention mechanism for new-created printed Chinese character recognition (CCR). Some studies \cite{zhang2020radical, cao2020zero} decompose characters into radicals following hierarchical structures or tree layouts, which model character recognition as an inference task. The above studies have made great progress in radical-based zero-shot character recognition. However, these methods cannot obtain satisfactory results when recognizing unseen or few-shot characters due to the lack of high-quality datasets with radical-level annotations, which affects their application in the real world.



\subsection{Radical Recognition}
In the past decades, researchers have also conducted studies on the radical-based recognition task because radicals are important components for composing characters and expressing semantics. Ma and Liu \cite{ma2008new} first attempt to separate radicals from characters by proposing an over-segmentation method that can handle left-right structure characters. Inspired by the hierarchical structures of different characters, Ma and Liu \cite{ma2009line} further explore a three-layer nested pre-segmentation method that could improve the radical segmentation performance for characters with left-right and top-down structures. Tan et al. \cite{tan2012radical} propose an affine sparse matrix factorization method for automatically extracting radicals from Chinese characters, addressing the poor alignment problem caused by the internal diversity of radicals. Deep neural networks are also applied for improving radical extraction, e.g., a deep residual network for detecting position-dependent radicals \cite{wang2017radical} and a deep CNN for analyzing radicals in Chinese characters \cite{yan2017rare}. To localize and recognize radicals in oracle images, Lin et al. \cite{lin2022radical} propose a radical extraction framework that introduces multi-scale features fusion and an attention mechanism to implicitly extract single radical features. Thus, high-quality benchmarks are required for such radical-based studies. 

\subsection{Character Image Datasets}
Several datasets have been published for the training and evaluation of OCR methods, e.g., HCL2000 \cite{zhang2009hcl2000} containing 3,755 simplified character categories, and IAHCC-UCAS2016 containing 350,621 handwritten character images. The large-scale ICDAR2013 competition database \cite{yin2013icdar} is the most commonly used dataset for evaluating the performance of handwritten Chinese character recognition (HCCR) methods. Researchers attempt to collect datasets by reproducing or painting characters, for instance, Oracle-20K \cite{guo2015building} contains 261 oracle bone categories with 20K handwritten samples, and HWOBC \cite{li2020hwobc} contains 83,245 handwritten samples. However, character reproduction is time-consuming and datasets with reproduced characters are limited when applied in practice since they change the actual data distribution. Thus, current studies focus more on collecting raw character images rather than reproducing characters, e.g., OBI125 \cite{yue2022dynamic} contains 30 categories, and the largest known OBI dataset OBC306 \cite{huang2019obc306} consists of 306 rubbing images. Note that all these datasets only have character-level annotations to support general image classification methods, which are unavailable for zero-shot character recognition methods or radical recognition methods. Therefore, in this study, we construct a large-scale ancient Chinese character image dataset with both radical-level and character-level annotations, which can support character/radical recognition tasks.


\section{Data Collection and Annotation}
\label{sec:3}
In this section, we first analyze the decomposition of characters that drives the motivation of our work. Then, we introduce the details of collecting ancient Chinese character images and producing radical-level annotations on the collected character images.

\begin{figure}[!t]
	\centering
	\setlength{\abovecaptionskip}{0pt}%
    \setlength{\belowcaptionskip}{-5pt}%
	\includegraphics[width=0.98\linewidth]{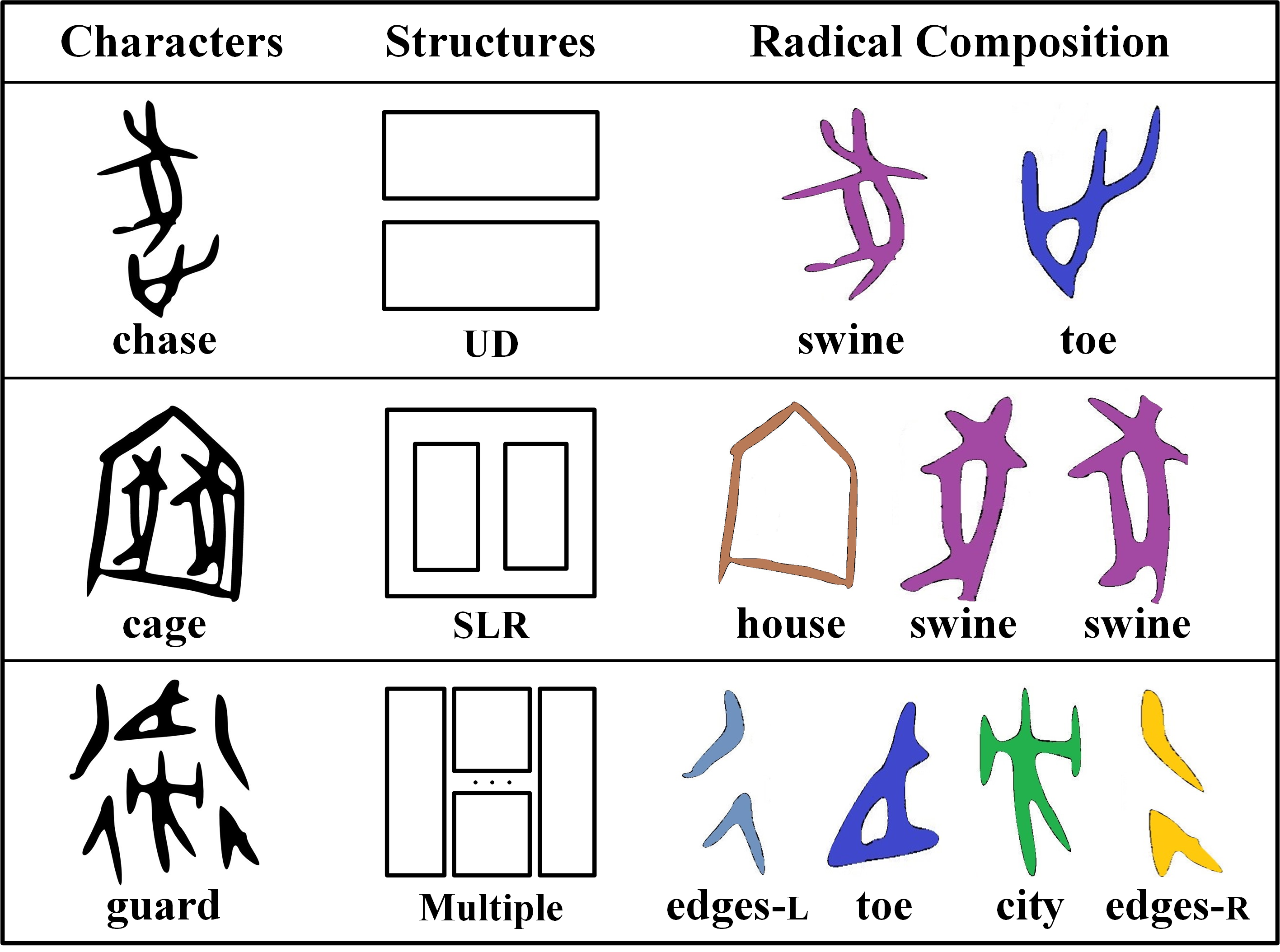}
	\caption{Examples of East Asian characters and their composing structures and radicals, radicals categories are distinguished by different colors. 
	\label{fig:Decomposition}}
\end{figure}

\subsection{Character Decomposition}
The characters of East Asian languages are symbolic characters, where using the orthography-based strategy in terms of radicals and structures is the most effective way to learn such characters \cite{shen2005investigation}.  There are two observations:
\begin{itemize}
    \item Characters can be decomposed into a set of radicals organized in specific structures according to orthography.
    \item Radicals are shared across different characters. The number of radicals is significantly less than that of characters.
\end{itemize}
Radicals are defined as graphic units of characters \cite{myers2016knowing}, where the number of radicals typically depends on the semantic complexity of the corresponding character. It is pointed out that radicals are the smallest components for expressing independent semantic information because they evolve from characters with simple meanings \cite{yeung2016orthographic}. The structures of characters also carry out character semantics since they are defined as the relative localization relations between constituent radicals. We list three examples in Fig.~\ref{fig:Decomposition} to demonstrate character decomposition. We can find the character \textit{chase} is presented by radicals \textit{swine} and \textit{toe} in an up-down (UD) structure, which means ``chasing swine by feet (toe)". The character \textit{cage} is composed of \textit{house} and \textit{swine} in a surrounding-left-right (SLR) structure, indicating ``the house to place animals”. Radicals in the character \textit{guard} indicate ``all edges of the city are protected", which represents a complex meaning by more radicals. Therefore, the discussion above shows that radical-level concepts, i.e., radicals and structures, are critical for representing the character from a linguistic perspective.  

Meanwhile, since radicals are simple-meaning components used to present complex semantics, radicals are shared across different characters. We can find examples in Figure ~\ref{fig:Decomposition}, where \textit{chase} and \textit{cage} share the radical \textit{swine}, and the \textit{chase} and \textit{guard} share the radical \textit{toe}. Such shareability allows a large number of characters can be presented by a few radicals, e.g., 6,763 common Chinese characters contain 485 radicals \cite{liu2011casia}, and 2,374 Korean characters contain only 68 radicals \cite{zatsepin2019fast}. 
Similarly, our proposed ACCID includes 2892 oracle bone characters which can be composed of 595 radicals. 
After decomposition, there are fewer categories of radicals and more samples in each category compared with that of characters. As a result, character recognition tasks benefit from decomposing characters into radical-level concepts when applying DL-based methods. Therefore, we aim to build a dataset with both character-level and radical-level concepts to improve character recognition performance.

\subsection{Character Image Collection}

\begin{figure}[!t]
	\centering
	\setlength{\abovecaptionskip}{0pt}%
    \setlength{\belowcaptionskip}{0pt}%
	\includegraphics[width=0.99\linewidth]{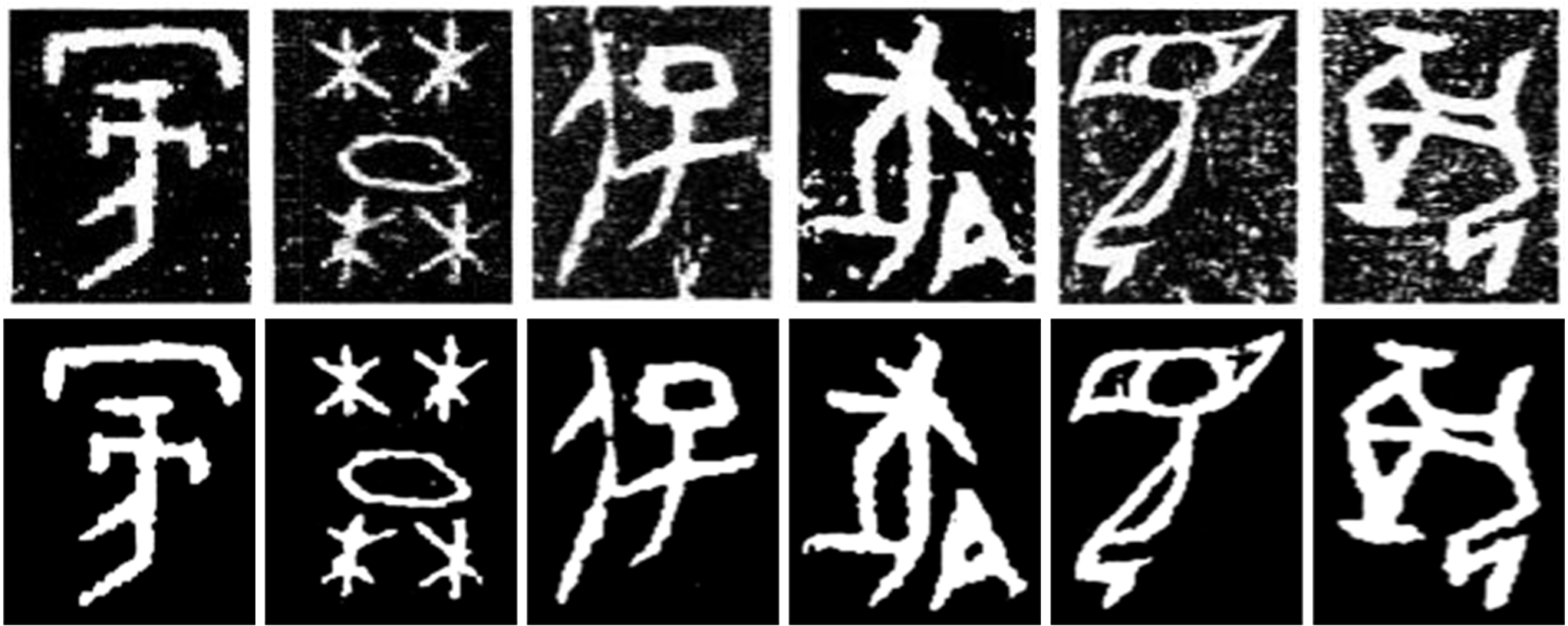}
    \caption{Examples of character images. Upper: raw character images with noise. Bottom: character images processed by restoration pipeline.	
    \label{denoising}}
\end{figure}

We start to build the dataset ACCID by collecting raw character images from an ancient character resource \cite{wu2012}. A character recognition tool \cite{redmon2018yolov3} is applied to locate and crop separate characters\footnote{Notice that all collected characters are OBIs.} automatically.
Meanwhile,  the categories of these character images $C_L$ can also be extracted automatically from the corresponding image captions. As shown in the first row of Figure~\ref{denoising}, we obtain raw character images with their character-level annotations. 

In order to guarantee the correctness of our collected data, we invited ten experts in the field of ancient Chinese characters to proofread these character-level annotations $C_L$. Specifically, we divided all images into five groups, each of which was proofread by two experts. The extracted character category will be retained if both experts agree with the current annotation, or modified if both experts agree with another annotation. In case of disagreement between experts, a third expert will be invited to arbitrate and determine the final annotation. Note that most of the ancient characters contain corrosion noise and friction traces, which limit the performance of character/radical recognition methods, and we can find examples of noisy raw images in Figure~\ref{denoising}. To solve this problem, we perform a pipeline for restoring character images, including noise removal \cite{shi2022charformer}, character completion, and super-resolution.
As shown in the second row of Figure~\ref{denoising}, the quality of character images is improved by applying the restoration pipeline. 


\begin{figure}[!t]
	\centering
	\setlength{\abovecaptionskip}{0pt}%
    \setlength{\belowcaptionskip}{0pt}%
	\includegraphics[width=0.99\linewidth]{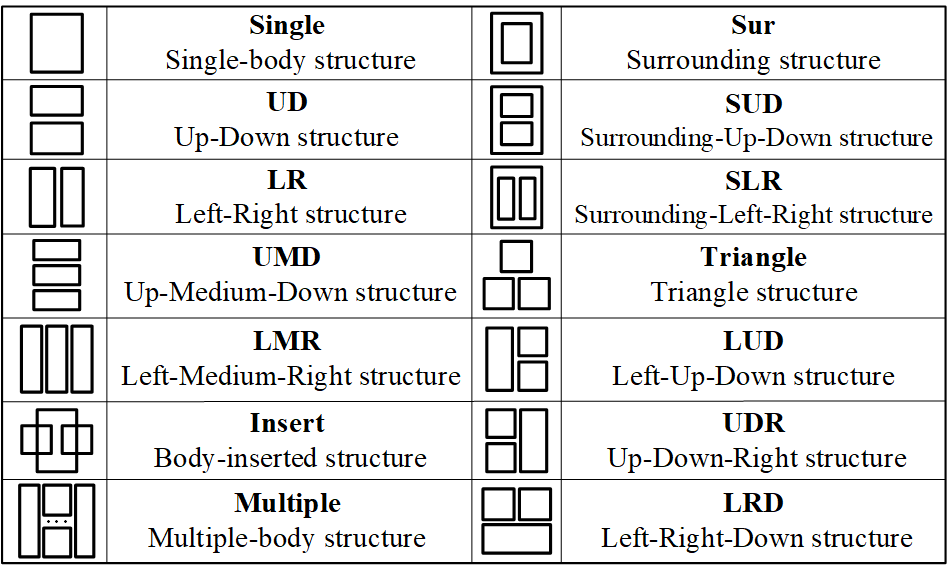}
	\caption{Pre-defined 14 categories of character structures. 
	\label{fig:structures}}
\end{figure}

\subsection{Radical-level Annotation}

In this section, we aim to annotate radical-level concepts, i.e., radicals and character structures, for our collected character images. Firstly, we obtain the concrete ancient Chinese character decomposition dictionary from a large Oracle knowledge graph (KG) ZiNet \cite{chi2022zinet} which contains 595 radicals and 14 structures for OBIs. Specifically, all character structures are pre-set by linguists,  as shown in Figure~\ref{fig:structures}. Then, we propose a pipeline for annotating radical-level concepts that applies both automatic and manual annotation to save the cost of human effort. 

\begin{figure}[!t]
	\centering
	\setlength{\abovecaptionskip}{0pt}%
    \setlength{\belowcaptionskip}{-5pt}%
	\includegraphics[width=0.99\linewidth]{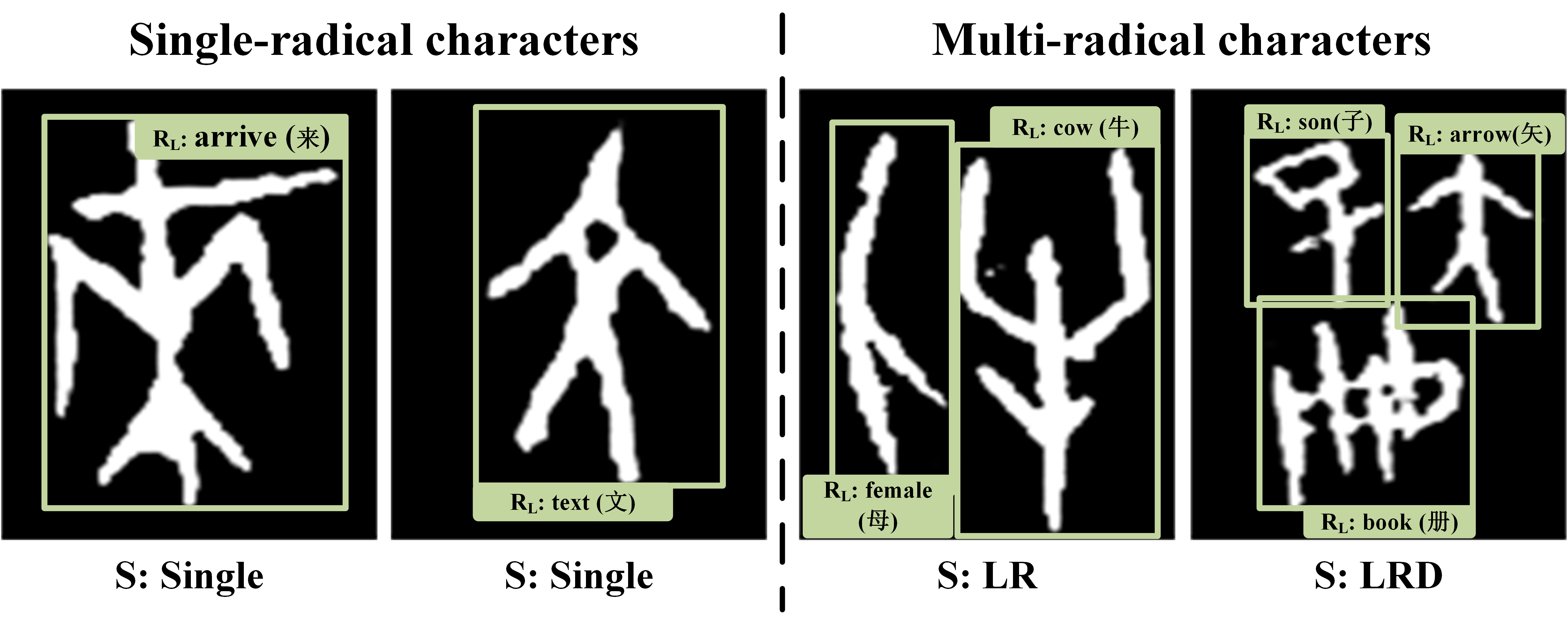}
	\caption{Demonstration of obtaining radical-level annotations in different cases.
	\label{fig:radical-level annotation}}
\end{figure}

As shown in Algorithm \ref{alg:0}, we introduce three kinds of annotations in the radical-level annotating pipeline $AnnRad(\cdot)$, including the label of radical category $R_L$, radical localization coordinates\footnote{We record the upper left and lower right corner coordinates to localize radicals.} $R_C$, and the label of structure category $S_L$. We demonstrate several examples of these annotations\footnote{We present both Chinese and English labels in $R_L$ for a better demonstration.} in Figure~\ref{fig:radical-level annotation}, where the coordinates $R_C$ are marked as boxes surrounding the radicals. Based on the radical number $k$ of a character obtained from the character decomposition dictionary by $getNum(\cdot)$, we consider two cases: \textit{single-radical characters} which contain one radical, and \textit{multi-radical characters} which contain two or more radicals. Single-radical characters are single-structured characters that are able to evolve into radicals. Thus, the category label $R_L$ of such radical can be directly regarded as its character category label $C_L$, and the structure category $S_L$ can be directly marked as ``Single". Meanwhile, we introduce an object detection method \cite{redmon2018yolov3} to localize single-radical characters for obtaining the coordinates $R_C$ of such radicals, as shown in the former two examples in Figure~\ref{fig:radical-level annotation}. In this way, we automatically annotate a large number of radicals, which significantly saves manual annotation effort.


Due to the diversity of radical composition and location in the \textit{multi-radical characters}, the knowledge of linguists is required to build a gold-standard dataset. We consider two steps for the annotation process. 
Firstly, we invited eight archaeological experts majoring in the field of ancient characters, to manually annotate the radical labels $R_L$, radical coordinates $R_C$ and the structural labels $S_L$. Specifically, $RadAnn_{E_i}(\cdot)$ and $StruAnn_{E_i}(\cdot)$ represent the annotation processes of the $i^{th}$ expert for the radical and structural information, respectively.
All \textit{multi-radical character} images are divided into four groups, each of which is annotated by two experts $E_1$ and $E_2$, as shown in the latter two examples in Figure~\ref{fig:radical-level annotation}. Secondly, two additional senior experts $SE$ are invited to proofread the annotations in the first step. If there is a disagreement between two experts, the senior expert will re-annotate the image and determine the final annotation, which helps to guarantee the quality of our proposed dataset.

\begin{algorithm}[!t] \small
\caption{ACCID radical-level annotation.
$R_L, R_C, S_L  = AnnRad(I)$}
\label{alg:0}
\begin{algorithmic}[1]
\REQUIRE Character images $I$;
\ENSURE Radical labels $R_L$, radical coordinates $R_C$, structure label $S_L$.

\STATE $k = getNum(I.C_L)$;
\IF {$k = 1$}\label{algln2}
    \STATE $R_L = I.C_L$;
    \STATE $R_C = getCoordinates(I)$;
    \STATE $S_L = Single$;
\ELSE
        \WHILE{$i \in range(k)$}
        \STATE $R_{Li1}, R_{Ci1} = RadAnn_{E_1}(I)$;   
        \STATE $R_{Li2}, R_{Ci2} = RadAnn_{E_2}(I)$;
        \IF {$R_{Li1} = R_{Li2}$ \& $R_{Ci1} = R_{Ci2}$}
            \STATE $R_{Li}, R_{Ci} = RadAnn_{E_1}(I)$;
        \ELSE
            \STATE $R_{Li}, R_{Ci} = RadAnn_{SE}(I)$;  
        \ENDIF 
        \STATE $R_L.append(R_{Li})$, $R_C.append(R_{Ci})$;
        \ENDWHILE
    \STATE $S_{L1} = StrAnn_{E_1}(I)$;
    \STATE $S_{L2} = StrAnn_{E_2}(I)$;
    \IF {$S_{L1} = S_{L2}$}
        \STATE $S_{L} = StruAnn_{E_1}(I)$;
    \ELSE
        \STATE $S_{L} = StruAnn_{SE}(I)$;
    \ENDIF
\ENDIF \\
\RETURN $R_{L}, R_{C}, S_{L}$; 
\end{algorithmic}
\end{algorithm}



\section{The Proposed Dataset ACCID}
\label{sec:4}
\subsection{Dataset Analysis and Demonstration}






The statistics of the ACCID are presented in Table \ref{tab:1}. The dataset contains 2,892 OBI character categories with a total of 15,085 images. The number of image samples for each character category (sample/cat.) varies from 1 to 33 due to the difference in character usage frequency and unearthed situations. 
According to the annotations by experts, we obtained 595 radical categories with a total of 28,143 samples, in which the number of samples per radical category ranges from 30 to 271. Each of the 14 pre-set structures contains a number of samples ranging from 236 to 5107. Several example images with both character-level and radical-level annotations can be found in Figure~\ref{fig:accid}. Note that all radical and structure categories have more than 30 samples. It is easy to find that the categories of radicals are less than those of characters (from 2,892 to 595), and character decomposition helps to significantly increase the average sample scale of each category (from 5.22 to 47.30). We can further say that our ACCID with radical-level data is appropriate for training DL models since it brings fewer categories and more samples per category, which makes it possible to improve the performance of radical and zero-shot character recognition. 

It should be mentioned that ACCID is a challenging dataset, because (1) unearthed oracle bone inscriptions usually have unavoidable and complex noise; (2) The data distribution of ACCID is similar to that of ancient characters in the real world, with limited sample size and significant long-tail effect.

\begin{table}[!t]
\centering
\setlength{\abovecaptionskip}{2pt}%
\setlength{\belowcaptionskip}{0pt}%
\resizebox{1 \linewidth}{!}{%
\begin{tabular}{@{}cccc@{}}
\toprule
                & \hspace{1em} Character \hspace{1em}        & \hspace{1em} Radical  \hspace{1em}    & \hspace{1em} Structure \hspace{1em}   \\\midrule
Category           & 2,892     & 595         & 14           \\ 
Sample             & 15,085    & 28,143         & 15,085        \\
Max sample/cat.    & 33       & 271         & 5107        \\
Min sample/cat.    & 1        & 30          & 236        \\\bottomrule
\end{tabular}}
\caption{The statistics of our proposed ACCID Dataset.}
\label{tab:1}
\end{table}

\begin{algorithm}[!t] \small
\caption{Splicing-based synthetic character image generation. 
$Set_{syn}  = GenImg(Set_{R}, Set_{S})$}
\label{alg:2}
\begin{algorithmic}[1]
\REQUIRE ~~\\ 
        Radical image set $Set_{R}$;
        character structure set $Set_{S}$.
\ENSURE 
        Synthetic character image set $Set_{syn}$.
        
\WHILE {$str \in Set_{S}$}
    \STATE $S_L = str$;
    \STATE $Num_R, Loc_R = getStructure(str)$;   

    \WHILE {$j \in range(n)$}
        
    \STATE $List_{tem} = getRadicals(Set_R, Num_R)$;
    \STATE $List_R = augmentImg(List_{tem})$;
        
    \STATE $R_L, R_C = getInfo(List_R)$;
    \STATE $I_{syn} = generateImg(List_R, Loc_R)$;
    \STATE $Syn_j.append(\{I_{syn}, R_L, R_C, S_L\})$
\ENDWHILE
    \STATE $Set_{syn}.append(Syn_j)$;
\ENDWHILE \\
\RETURN $Set_{syn}$; 
\end{algorithmic}
\end{algorithm}

\begin{figure}[!t]
	\centering
	\setlength{\abovecaptionskip}{0pt}%
    \setlength{\belowcaptionskip}{0pt}%
	\includegraphics[width=1\linewidth]{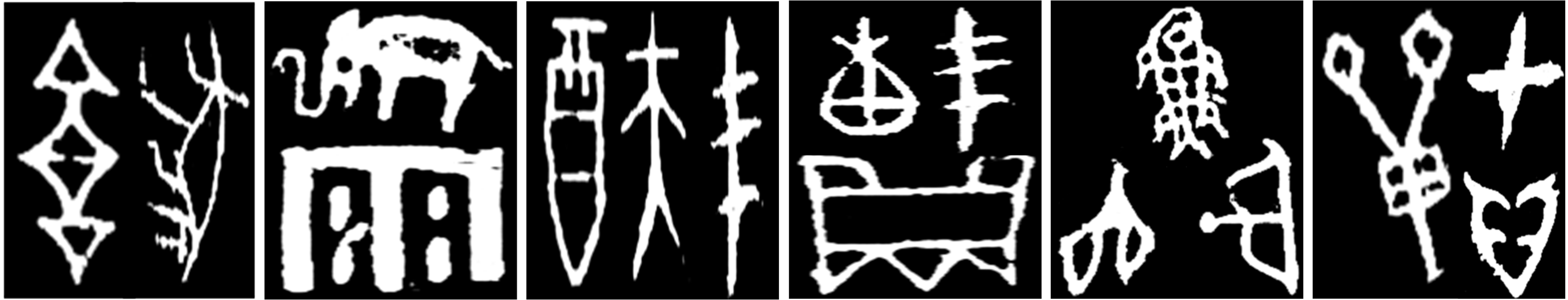}
	\caption{Examples of synthetic images generated by splicing-based synthetic character strategy in ACCID$_{Syn}$. \label{fig:pj1}}
\end{figure}

\subsection{Splicing-based Synthetic Character Strategy}

Researchers highlight that the number of OBI image samples is limited due to the scarcity of unearthed OBIs \cite{huang2019obc306}, which results in a shortage of training data in existing OBI datasets \cite{yue2022dynamic, lin2022radical}. To provide adequate training samples, we propose a splicing-based strategy to generate synthesized images based on radical-level concepts. The detailed process is shown in Algorithm \ref{alg:2}, where the main purpose is to reuse radical-level annotations to generate more ``characters" images for model training without additional human efforts.

According to the annotated radical categories and coordinates, we collect all radical images $Set_{R}$ by cropping the raw character images. We plan to generate $n$ synthetic images for each structure in $Set_{S}$ that contains 13 pre-defined structures except $Single$. For each structure, we identified the number of contained radicals $Num_R$ and relative localization $Loc_R$ between radicals using $getStructure(\cdot)$. Then, $getRadicals(\cdot)$ randomly selects a list of radical $List_{tem}$ from  $Set_{R}$ to compose the $j-th$ synthetic character. We propose a pre-processing pool, including \textit{zooming in}, \textit{zooming out}, \textit{rotating}, \textit{distorting}, and \textit{padding}, where these operations are randomly selected to augment radicals in $List_{tem}$ using $augmentImg(\cdot)$. According to $Loc_R$ and augmented radicals $List_R$, $generateImg(\cdot)$ splices all radicals in $List_{tem}$ as a character image. Meanwhile, we also record the corresponding structure category $S_L$, the selected radical category $R_L$ and their localization coordinates $R_C$, along with the generated synthetic character images $I_{syn}$. Finally, we obtain a synthetic character set $Set_{syn}$. We demonstrate some examples of synthetic character images in Figure~\ref{fig:pj1} It is worth mentioning that this splicing-based character synthesis strategy is also adaptive for other East Asian languages containing radicals, such as Chinese, Korean, and Japanese, which provides a feasible means for data augmentation in datasets of such languages. Overall, we set $n = 3,600$ and generate a total of 46,800 synthetic characters in the character image set ACCID$_{Syn}$.

\section{Baseline Method}
\label{sec:5}

\begin{figure}[!t]
	\centering
		\setlength{\abovecaptionskip}{2pt}%
    \setlength{\belowcaptionskip}{0pt}%
	\includegraphics[width=1\linewidth]{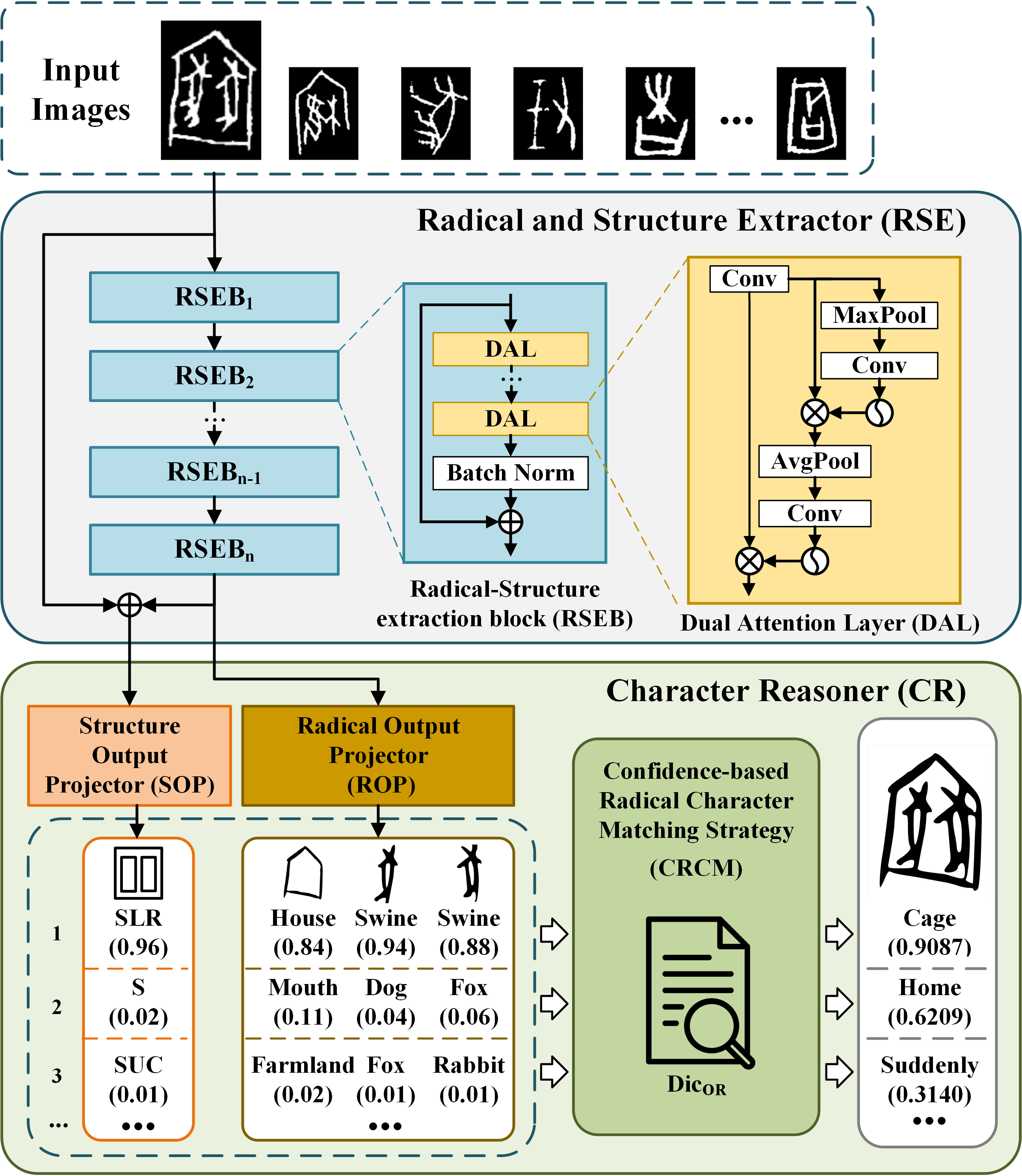}
	\caption{The overall architecture of our baseline method. 
	\label{fig:baseline}}
 \vspace{-0.2cm}
\end{figure}

We propose a trainable baseline for zero-shot OCR by character decomposition and reorganization. Note that we are not exploring an optimal model with maximized performance. Instead, we are interested in learning about the usability of the proposed ACCID on zero-shot OCR and identifying venues for future research. The baseline is a two-step method, and its architecture is illustrated in Figure~\ref{fig:baseline}. First, the radical and structure extractor (RSE) aims to extract the structure and radicals from the input character image. Then, the target character category is recognized by the character reasoner based on the extracted information.  


\subsection{Radical and Structure Extractor}
In the RSE, a set of Radical-Structure Extraction Blocks (RSEB) are designed as the backbone network to extract deep features from the input image, where each RSEB is composed of several dual attention layer (DAL) \cite{diao2023rzcr} and a batch normalization layer, as shown in the blue blocks in Figure~\ref{fig:baseline}. The DAL aims to issue overlapping and unclear boundaries between radicals, where attention weights are obtained from two computations, as shown in the yellow blocks in Figure~\ref{fig:baseline}.
Due to radicals and structure in a character being associated and both containing semantic information, we engage to make RSEB learn both radical and structure features to extract them efficiently. As a result, RESBs output the deep feature $F_r$ that contains semantic information to perform radical detection and structural relation extraction in parallel. 

Two output projectors, radical output projector (ROP) and structure output projector (SOP), are designed to simultaneously recognize structures and radicals, where ROP constrains the extractor training by predicting radical categories and their localization, and SOP constrains RSEB to simultaneously learn about structure information that also facilitates radical extraction by ROP. 
The output projector ROP consists of two convolutional layers and an FC layer whose size is $K{\times} K{\times} M{\times}(n_r{+}n_c)$, where $K{\times} K$ refers to the number of divided grids of an input character image, $M$ represents the number of anchor boxes in each grid
, $n_r$ is the number of radical categories in the datasets, and $n_c$ records the coordinates of the radical location ($x, y, w, h$) and the confidence of radical detection, thus $n_c {=} 5$. Note that we set $K {=} 13$ and $M {=} 3$ according to experimental results.
Parallelly, we apply another output projector SOP to predict the structure of characters, where both shallow features (the output features of RESB$_1$) and deep features $F_r$ (containing radical location information) are exploited to capture the global and local structural information. The SOP consists of five convolutional layers and an FC layer to further handle the concatenated shallow-deep features.

\subsection{Character Reasoner}

\begin{algorithm}[!t] \small
\caption{Confidence-based Radical Character Matching. \\
$PC = CRCM(Dic_{OR}, R, S)$}
\label{alg:3}

\begin{algorithmic}[1] 

\REQUIRE ~~\\ 
        Character Dictionary $Dic_{OR}$;
        Candidate radicals $R$; \\
        Candidate structures  $S$.\\
\ENSURE ~~\\ 
Character predictions with confidence $PC$.\\

\STATE $P_{R}.append( \frac{1}{n}\sum_{i=1}^{n} R^i_j.conf$);   \label{line:1}
\STATE $P_{S}.append(S_k.conf)$; 

\FOR{each $j, k \in TopConf(P_{R_j}, P_{S_k}, t)$}
\STATE $List = \{R^1_j, R^2_j, ..., R^n_j, S_k\}$; 
\IF {$c =  searchDic (List, Dic_{OR})$ }
\STATE $p_{c} = \theta P_{R_j} +  (1 - \theta)P_{S_k}$; \\   \label{line:7}
\STATE $PC.add(C, p_{c})$  \\
\ENDIF
\ENDFOR

\RETURN $maxSort(PC)$ \\
\end{algorithmic}
\end{algorithm}

After obtaining the radicals and candidate characters from the RSE, the target character is recognized by utilizing a character dictionary in the character reasoner (CR). The dictionary is extracted from a public oracle knowledge graph $ZiNet$ \cite{chi2022zinet}, which stores character categories and their corresponding decomposition information, including radicals and structures. To enhance the matching between the characters and radical sets, we propose a confidence-based radical character matching strategy, called CRCM,  aiming to make full use of the confidence obtained from the RSE. The proposed CRCM is denoted as $CRCM(\cdot)$, whose inputs are the oracle dictionary $Dic_{OR}$, candidate radicals ($R$), and candidate structures ($S$), and output is the candidate characters and their confidence.

The matching process is shown in algorithm \ref{alg:3}. First, we calculate the average confidence $P_{R}$ of the possible candidate radical sets output by ROP, as shown in Line \ref{line:1}, where $n$ is the number of radicals recognized by RSE, $R^i_j.conf$ is the prediction confidence $conf$ of the $j^{th}$ candidate radical at the $i^{th}$ location in a character. The structure confidence $P_S$ includes the prediction confidence $S_k.conf$ of the $k$ candidate structure output by SOP. We combine the Top-$t$ candidate radical sets and structures respectively and match the characters in $Dic_{OR}$ through $searchDic(\cdot)$, to get the candidate character $C$. In the experiments, we set $t = 5$. The confidence $p_c$ of $C$ is calculated by $P_{R_j}$ and $P_{S_k}$ in Line \ref{line:7}, where $\theta{=}0.7$. The matched candidate character $C$ and the corresponding confidences $p_c$ are stored in character prediction $PC$. We output the sorted $PC$ as the recognition results.

The proposed character-matching method comprehensively considers the extracted character information from the baseline network, which effectively alleviates the low-precision character reasoning issue caused by hard-matching strategies. Overall, the baseline is a zero-shot method that is able to recognize unseen character categories by the soft matching strategy.

\section{Evaluation}
\label{sec:6}

To learn about the quality and challenges of the proposed dataset, we designed a series of metrics and experiments to evaluate ACCID from various perspectives. We also apply the proposed trainable baseline for zero-shot character recognition by character decomposition and reorganization, as we introduced in Sec.5. Note that we are not exploring an optimal model with maximized accuracy, instead, we are interested in learning about the usability and challenges of the proposed ACCID on the zero-shot character recognition tasks and identifying venues for future research.

\subsection{Dataset Quality Assessment}


\noindent \textbf{Inter-class Agreements.}
To evaluate the quality of ACCID annotations, we introduce the widely-used reliability coefficients \textit{Krippendorff's alpha} \cite{k-alpha} to evaluate the inter-class agreement. According to the annotation process, the character-level annotation is completed by five groups of experts, the radical-level annotation is completed by four groups of experts, and each image is annotated by at least two experts. Thus, we record the average within-group \textit{Krippendorff's alpha} ($Avg.~\alpha$), as shown in Table~\ref{tab:2}, the $Avg.~\alpha$ of character category labels $C_L$ achieve $0.9852$, which means near-perfect agreement. Radical-level labels, including radical category labels $R_L$, radical coordinate $R_C$, and structure category labels  $S_L$ get $0.9508$, $0.9295$, and $0.9890$ for $Avg.~\alpha$, respectively, also achieve excellent agreement. The high coefficient shows that the annotators were properly chosen and did not introduce bias, which validates the reliability of ACCID.

\begin{table}[!t]
\centering
  \setlength{\abovecaptionskip}{0pt}%
  \setlength{\belowcaptionskip}{0pt}%
  \resizebox{0.9\linewidth}{!}{%
\begin{tabular}{@{}ccccc@{}}
\toprule
              & $C_L$ \hspace{2em}  & $R_L$  \hspace{2em}  & $R_C$ \hspace{2em}  & $S_L$ \hspace{2em}  \\ \midrule
$Avg.~\alpha$ \hspace{2em} & 0.9852 \hspace{2em} & 0.9508  \hspace{2em}   & 0.9295 \hspace{2em}  & 0.9890 \hspace{2em} 
 \\ \bottomrule
\end{tabular}}
\caption{The inter-class agreements between annotators for different-level labels in ACCID. (Krippendorff's Alpha)}
\vspace{-0.4cm}
\label{tab:2}
\end{table}




\noindent \textbf{Evaluation on Different-level Annotations.}
To comprehensively evaluate the quality of ACCID, we design three sets of experiments for four kinds of labels belonging to the character level and the radical level.

\noindent \textbf{\textit{Experiments setup.}} The same settings are applied for all experiments. The resolution of the input images is $256 {\times} 256$, and data augmentation strategies including translation, rotation, and scaling are exploited during the training. 80\% of data is used as training set and 20\% is the test set. All experiments are performed with Adadelta optimization with hyperparameters set to $\rho {=} 0.95$ and $\varepsilon {=} 10^{-6}$. 
The initial learning rate is 1e-4, and the cross-entropy loss is applied in classification.

\noindent \textbf{\textit{Character level annotations.}} To evaluate character category labels $C_L$ in ACCID, we introduce three generic image classification networks and two state-of-the-art OCR methods for experiments, which we collectively call character-based methods. As shown in the first column of Table~\ref{tab:3},  all methods achieved more than 42\% accuracy in character recognition. Note that the performance is limited by the difficulty of the real-world dataset rather than incorrect annotations, since the reliability of ACCID has been verified by the great agreement in Table~\ref{tab:2}. ACCID constructed in real scenarios suffers from unbalanced and insufficient training samples of character categories, which brings challenges and opportunities for future research, especially for zero-shot character recognition.


\noindent \textbf{\textit{Radical level annotations.}}
The radical-level annotation consists of three parts, in which the structural category label $S_L$ is verified by the above-mentioned multi-classification methods, while the radical category $R_L$ and the radical location $R_C$ are introduced into object detection models for evaluation. As shown in Table~\ref{tab:3}, structure classification achieves good performance on all methods, verifying the reliability of $S_L$. To evaluate $R_L$ and $R_C$, we introduce the metric $AP_{50}$, namely, the mean average precision ($mAP$) of radical categories when the Intersection Over Union (IOU) between true and predicted $R_C$ is 50\%. The third column of Table~\ref{tab:3} gives the performance of radical detection. The impressive $AP_{50}$ results demonstrate the availability of $R_L$ and $R_C$ and the high quality of radical-level annotations.

\begin{table}[!t]
\centering
  \setlength{\abovecaptionskip}{0pt}%
  \setlength{\belowcaptionskip}{0pt}%
  \resizebox{1\linewidth}{!}{%
\begin{tabular}{@{}ccc|cc@{}}
\toprule
Methods                                 & Character Acc.     & Structure Acc.              & Methods                                       & Radical $AP_{50}$ \\ \midrule
VGG16 \cite{simonyan2014very}           & 42.32\%            & 73.72\%       
& Faster R-CNN \cite{ren2015faster}             & 63.76\%     \\
ResNet \cite{he2016deep}                & 46.18\%            & 75.69\%       
& SSD \cite{liu2016ssd}                         & 57.45\%     \\
GoogleNet \cite{szegedy2013intriguing}  & 43.27\%            & 77.21\%       
& YOLOv3 \cite{redmon2018yolov3}                & 64.02\%     \\
DirectMap\cite{zhang2017online}         & 52.36\%            & 76.63\%           
& YOLOv4 \cite{bochkovskiy2020yolov4}           & 56.68\%     \\
M-RBC + IR\cite{yang2017improving}      & 53.74\%            & 77.94\%          
& RetinaNet \cite{lin2017focal}                 & 66.23\%     \\
\bottomrule
\end{tabular}}
\caption{Accuracy of character, radical and structure recognition on ACCID.}
\label{tab:3}
\vspace{-0.3cm}
\end{table}


\subsection{Results of ACCID}

\begin{table}[!t]
\centering
\setlength{\abovecaptionskip}{0pt}%
\setlength{\belowcaptionskip}{0pt}%
	\resizebox{0.8\linewidth}{!}{%
\begin{tabular}{@{}lcccc@{}}
\toprule


Method & Top-1 & Top-3 & Top-5 & Cat$_{Avg}$  \\ \midrule

AlexNet \cite{krizhevsky2012imagenet} & 26.93\% & 36.45\% & 40.03\%  & 21.74\% \\
VGG16 \cite{simonyan2014very}        & 27.75\% & 38.12\% & 41.53\% &  20.38\%  \\
HCCR-GoogLeNet \cite{zhong2015high}  & 28.52\% & 36.75\% & 39.86\%    &  18.81\% \\
DropSample-DCNN\cite{yang2016dropsample} & 29.19\% & 39.27\%  & 42.03\%  & 19.59\% \\
ResNet \cite{he2016deep}             & 28.50\% & 33.02\%  & 40.66\%  &  21.98\%  \\
DenseNet \cite{huang2017densely}     & 27.85\% & 38.63\% & 47.48\% & 19.20\% \\ 
DirectMap\cite{zhang2017online}      & 30.48\% & {\color[HTML]{0070C0}44.89\%} & {\color[HTML]{0070C0}54.72\%} & 23.59\%  \\   
M-RBC + IR\cite{yang2017improving}   & 30.53\% & 42.74\% & 49.32\% & 20.72\%  \\ 
\midrule

RAN \cite{zhang2018radical}          & 35.37\% & - & -  & 32.48\% \\
DenseRAN \cite{wang2018denseran}     & 36.02\% & - & -  & 32.16\%   \\                     
FewshotRAN \cite{wang2019radical}    & 33.31\% & - & -  & 30.90\%  \\              
HDE-Net \cite{cao2020zero}           & {\color[HTML]{0070C0}36.79\%} & - & - & {\color[HTML]{0070C0}33.10\%}  \\                    
Stroke-to-Character \cite{chen2021zero} & 27.30\% & - & - & 20.09\% \\ 

Baseline (Ours) & {\color[HTML]{FF0000}60.28\%} & {\color[HTML]{FF0000}69.74\%} & {\color[HTML]{FF0000}72.66\%}  & {\color[HTML]{FF0000}58.17\%}  \\ \bottomrule
\end{tabular}}
\caption{Quantitative comparisons with state-of-the-art methods on ACCID. The best and second-best results are highlighted in {\color[HTML]{FF0000} red} and {\color[HTML]{0070C0} blue} colors, respectively.}
\label{tab:6}
\vspace{-0.5cm}
\end{table}

We apply the proposed baseline method on ACCID for zero-shot character recognition and compare it with state-of-the-art character recognition methods. The results are shown in Table \ref{tab:6}. We output Top-n prediction by confidence to present the average classification accuracy based on samples. Considering a few categories with more samples are not enough to reflect the overall recognition performance in an unbalanced dataset, we also calculate the average accuracy for each category and then average over all categories, i.e., Cat$_{Avg}$. Note that in Table \ref{tab:6}, character-based OCR methods are presented in the former rows, and the latter are zero-shot OCR methods. In ACCID, we select 80\% of the samples in each character category as the training set and the remaining as the test set. Note that categories containing only one sample are not included in this experiment since character-based recognition methods are not able to train on these categories. We can find the proposed baseline method significantly outperforms all character-based OCR methods and zero-shot methods on Top-n accuracy and Cat$_{Avg}$, which demonstrates the proposed zero-shot character recognition effectively utilized radical-level annotations in ACCID and show the superiority.
Character-based OCR methods obtained lower Cat$_{Avg}$ than zero-shot character OCR methods means that these methods poorly perform on categories with few samples. The main reason is that an insufficient amount of training data limits their performance. In contrast, the operation that decomposes characters into elements in zero-shot recognition methods brings an increasing number of training samples and a decrease in training categories, which alleviates the few-sample issue. 
The proposed baseline method performs better than all others on metric Cat$_{Avg}$, which proves our method is less influenced by categories with different numbers of samples during recognition.

\begin{table}[!t]
\centering
\setlength{\abovecaptionskip}{0pt}%
\setlength{\belowcaptionskip}{0pt}%
\resizebox{0.70\linewidth}{!}{
\begin{tabular}{@{}lccc@{}}
\toprule
Method & c-500 & c-1000 & c-1205  \\ \midrule
DenseRAN \cite{wang2018denseran} & 5.28\% & 10.67\% & 11.58\%      \\
HDE-Net \cite{cao2020zero}       & 7.12\% & 9.76\% & 10.51\%        \\
Baseline (Ours)   & \textbf{38.56\%} & \textbf{51.80\%} & \textbf{53.74\%}  \\ \bottomrule
\end{tabular}}
\caption{Comparisons with zero-shot character recognition methods on ACCID.}
\label{tab:7}
\vspace{-0.6cm}
\end{table}

\subsection{Results of Zero-shot Character Recognition}

We conduct an experiment to demonstrate the effectiveness of the baseline method on zero-shot character recognition. Since character-based methods cannot recognize unseen character categories, only zero-shot recognition methods are included in this experiment. We applied $n$ comparison group of training sets, fixed the test categories in each experiment, and gradually increased the number of training categories. Specifically, we select samples of the first  $n$ character categories from ACCID as seen categories for training, where  $n {\in} \{500, 1000, ...\}$. Then we select the last 800 character categories from ACCID as unseen categories for testing. The results are shown in Table \ref{tab:7}.

We can find that our proposed baseline method significantly outperforms other sequence-based zero-shot methods in each set of datasets, which further proves the superiority of the recognition strategy utilizing radical-level information for character reasoning. More specifically, the proposed baseline method benefits from the confidence-based radical character matching strategy, which results in the possibility of correct character recognition even with incorrect Top-1 radical prediction. As a result, we can conclude that the proposed zero-shot character recognition baseline is effective for the proposed significant few-sample dataset.

\section{Conclusion}
In this paper, we introduce ACCID, an ancient Chinese character image set with multiple types of annotations, including character categories, structures, radical categories, and radical locations. ACCID aims to complement a benchmark for zero-shot character recognition tasks. Meanwhile, we propose a baseline method to analyze the usability and challenges of ACCID. 
We evaluate ACCID from various perspectives, and experimental results show that annotations in ACCID are high-quality and reliable. 
Overall, as a newly proposed benchmark, we believe ACCID moves us toward zero-shot character recognition, which is valuable to deal with unbalanced samples in real-world application scenarios.

\begin{acks}
This work is supported by the ``Paleography and Chinese Civilization Inheritance and Development Program" Collaborative Innovation Platform (No. G3829), Jilin University Philosophy and Social Science Research Innovation Team Youth Project (No. 2023QNTD02), and the Department of Science and Technology of Jilin Province, China (No. 20230201086GX).
\end{acks}

\bibliographystyle{ACM-Reference-Format}
\balance
\bibliography{accid}


\end{document}